\documentclass[runningheads]{llncs}

\usepackage{iciap}

\usepackage{iciapabbrv}

\usepackage{graphicx}
\usepackage{booktabs}
\usepackage{svg}

\usepackage[accsupp]{axessibility}  

\usepackage{hyperref}

\usepackage{orcidlink}

\begin{document}

\title{Mamba-OTR: a Mamba-based Solution for Online Take and Release Detection from Untrimmed Egocentric Video} 

\titlerunning{Mamba-OTR}

\author{Alessandro Sebastiano Catinello \and
Giovanni Maria Farinella \and
Antonino Furnari}

\authorrunning{A.~Catinello et al.}

\institute{Department of Mathematics and Computer Science - University of Catania, Italy
\email{ale.catinello.c@gmail.com}\\
\email{\{giovanni.farinella, antonino.furnari\}@unict.it}}

\maketitle

\begin{abstract}
This work tackles the problem of Online detection of ``Take'' and ``Release'' (OTR) of an object in untrimmed egocentric videos. This task is challenging due to severe label imbalance, with temporally sparse positive annotations, and the need for  precise temporal predictions. Furthermore, methods need to be computationally efficient in order to be deployed in real-world online settings. To address these challenges, we propose Mamba-OTR, a model based on the Mamba architecture. Mamba-OTR is designed to exploit temporal recurrence during inference while being trained on short video clips. To address label imbalance, our training pipeline incorporates the focal loss and a novel regularization scheme that aligns model predictions with the evaluation metric. Extensive experiments on EPIC-KITCHENS-100, the comparisons with transformer-based approach, and the evaluation of different training and test schemes demonstrate the superiority of Mamba-OTR in both accuracy and efficiency. These finding are particularly evident when evaluating full-length videos or high frame-rate sequences, even when trained on short video snippets for computational convenience.
The proposed Mamba-OTR achieves a noteworthy mp-mAP of $45.48$ when operating in a sliding-window fashion, and $43.35$ in streaming mode, versus the $20.32$ of a vanilla transformer and $25.16$ of a vanilla Mamba, thus providing a strong baseline for OTR. We will publicly release the source code of Mamba-OTR to support future research.

\end{abstract}

\section{Introduction}
\label{sec:intro}

Wearable devices provided with cameras are able to capture visual information from a first-person perspective, enabling the development of personalized, context-aware assistive technologies to support user daily activities~\cite{egovisionoutlook}. A key requirement for such systems is the ability to detect fine-grained, atomic actions—such as \textit{take} and \textit{release} of an object—which are essential for downstream tasks like intention prediction, object interaction tracking, and anomaly detection during goal-directed human behavior.
\begin{figure}
    \centering
    \includegraphics[width=\linewidth]{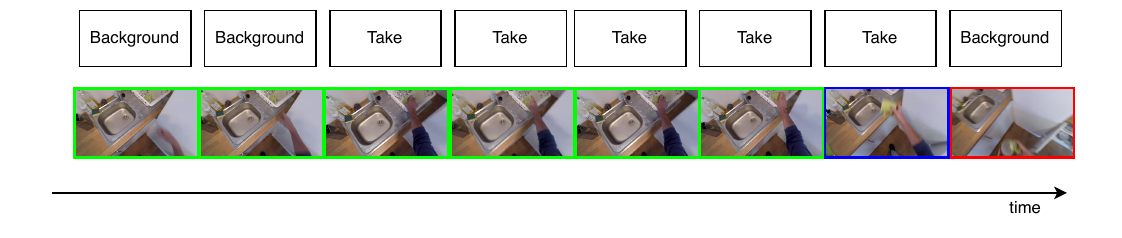}
    \caption{Algorithms are tasked to process the video online and output a single prediction corresponding to the last frame of the take/release action (blue frame), while avoiding predictions for background (red frame) or any other frame (green ones).}
    \label{fig:ODAE}
\end{figure}
To address this challenge, action recognition algorithms must operate on streaming video data in an online fashion while maintaining temporal coherence by emitting a single, unambiguous prediction per action instance. We refer to this task as ``Online Take-Release detection'' (OTR). Previous work has explored various strategies, including detecting the \textit{starting} frame of an action~\cite{shou2018online} and identifying \textit{contact} frames~\cite{scavo2023quasi}. However, recent findings suggest that predicting the \textit{ending} frame of an action yields superior performance, as it reduces the uncertainty caused by partial observations and premature predictions~\cite{catinello863online}, while inducing a small delay in the predictions, which is acceptable for most applications. \cref{fig:ODAE} illustrates this setup, originally discussed in~\cite{catinello863online}. 
Despite the advantages of this formulation, OTR still faces multiple challenges, notably, the extreme class imbalance between positives (last frame of a take/release action) and negatives (any other frame), the need to suppress multiple detections, and the requirement of online and computationally efficient processing in order to support deployment to real-world scenarios.

We propose a novel approach based on the Mamba architecture~\cite{gu2023mamba}. Our method uses the focal loss during training and incorporates tailored regularization techniques that align the model's behavior with the evaluation metric, encouraging precise and temporally consistent predictions. We evaluated our model on the EPIC-KITCHENS-100 dataset, focusing exclusively on the \textit{take} and \textit{release} verbs, and conduct extensive experiments across various architectural configurations.
The proposed Mamba-OTR has been compared with respect to different approaches based on Tranformers and Mamba-based models, showing the advantages of each introduced optimization.
Our results demonstrate that, although the task remains challenging, it becomes significantly more tractable when approached with an appropriate training strategy. Furthermore, we show that Mamba’s inherent recurrence enables efficient training on short clips while allowing inference over full-length videos. This training-inference decoupling leads to substantial improvements in inference speed and enhances predictive performance.
Our optimized Mamba-OTR model achieves an mp-mAP of $51.76$ and a mean inference time of $0.14s$, versus the mp-mAP of $20.32$ and the inference time of $0.28s$ of a standard Transformer module.

In sum, our main contributions are as follows: 1) A training pipeline incorporating regularization strategies to address severe annotation imbalance of OTR; 2) Comprehensive benchmarking against existing Transformer- and Mamba-based methods; 3) The Mamba-OTR model, which is a strong baseline to support research in this area. We plan to release our code.

\section{Related Work}
\label{sec:related_work}

\paragraph{\textbf{{Online Detection of Action Start/End}}}
The considered OTR problem is closely related to previous investigations aiming to detect action start or end frames in an online fashion.
The Online Detection of Action Start (ODAS) task~\cite{shou2018online} aims to identify the exact moment an action begins in a streaming setting, emphasizing low latency and temporal precision. Several methods have approached this challenge using third-person video with 3D convolutional networks~\cite{shou2018online}, LSTM-based models with reinforcement learning~\cite{gao2019startnet}, and weak supervision~\cite{gao2021woad}. 
More recently, quasi-online strategies tailored explicitly for OTR have relaxed strict causality by introducing short buffer windows~\cite{scavo2023quasi}.
While similar in evaluation and setup, the Online Detection of Action End (ODAE) task~\cite{catinello863online} proposes predicting the frame where the action concludes, leveraging the full temporal extent of the action and reducing the ambiguity present in early-stage predictions. For applications involving real-time monitoring, end-point predictions provide a sufficiently clear signal for downstream tasks.

In this work, we consider the ODAE formulation proposed in~\cite{catinello863online} to address the OTR problem, given its reduced ambiguity and practical relevance.

\paragraph{\textbf{{Deep Learning Architectures for Online Video Processing}}}
Research in online video processing aims to balance the temporal accuracy of detections with computational efficiency. Transformer-based models, such as OadTR~\cite{wang2021oadtr}, TeSTra~\cite{shou2018online} or LSTR~\cite{xu2021long} have been proposed. In particular, these two last methods employ dual-memory mechanisms to integrate both short- and long-term temporal dependencies, but their limited scalability poses challenges in real-time applications. To address these limitations, alternative architectures based directly on recurrent neural networks~\cite{an2023miniroad, xu2019temporal} or Mamba~\cite{gu2023mamba, chen2024video} architectures have been proposed. These latter models performs sequence modeling in linear time using selective state spaces, offering competitive or superior performance compared to Transformers while reducing computational complexity in both time and space.

In this work, we adopt the Mamba architecture due to its ability to parallelize training and operate in a recurrent manner at inference time, making it particularly well-suited for efficient online video processing.

\paragraph{\textbf{{Class imbalance}}}
Class imbalance is a persistent challenge in deep learning, particularly in object detection, where background regions vastly outnumber foreground objects. Solutions like \textit{focal loss}~\cite{lin2017focal} prioritize hard examples, while \textit{class-balanced loss}~\cite{cui2019class} and \textit{LDAM loss}~\cite{cao2019learning} adjust for skewed label distributions via re-weighting and margin tuning.
Although developed for image-based tasks, these methods are applicable to temporal problems such as action detection, where background frames dominate and relevant actions are short. 

In this paper, we show that adapting imbalance-aware losses is crucial for improving performance in online action detection.




\section{Problem Definition and Evaluation Metrics}
\label{sec:problem_definition_evaluation}

Following the work in~\cite{catinello863online}, we define OTR as the task of identifying whether the current frame at time \( t' \) marks the end of a \textit{take} or \textit{release} action, using only video frames observed up to \( t' \). Each ground truth action is defined as \( a = (c, t) \), where \( c \) is the action class and \( t \) is the end timestamp. Predictions are expressed as tuples \( \hat{a} = (\hat{c}, \hat{t}, s) \), including the predicted class $\hat{c}$, predicted timestamp $\hat{t}$, and confidence score $s$.
To evaluate our models, we adopt the point-level mean Average Precision (p-mAP) metric, as introduced in~\cite{shou2018online}. A predicted action \(\hat{a} = (\hat{c}, \hat{t}, s)\) is matched to a ground truth action \(a = (c, t)\) if the following conditions are met:
1) The predicted and ground truth action classes match \((\hat{c} = c)\); 2) The temporal offset \(\delta = |\hat{t} - t|\) is less than or equal to a threshold~\(\phi\).
Matching is performed greedily based on descending confidence scores. Each prediction and each ground truth can be matched at most once. Matched predictions are counted as true positives; unmatched predictions are false positives, and unmatched ground truth actions are false negatives. Take/Release mAP is computed based on these true and false positives, considering confidence scores $s$.
To account for different levels of temporal precision, we report the mean point-level mAP (mp-mAP), computed by averaging the p-mAP values over a range of temporal thresholds~\(\phi\) ranging from 1 to 10 seconds (with 1-second intervals).

\section{Architecture and Techniques}
The proposed Mamba-OTR is obtained by combining a streamlined Mamba architecture with the focal loss and different regularization techniques. For comparison, we assess the effect of the proposed techniques on both Transformer-based and Mamba-based architectures. In the following, we first describe base architectures, then provide details on the training loss and proposed regularization techniques. 

\begin{figure}[t]
    \centering
    \includegraphics[width=\linewidth]{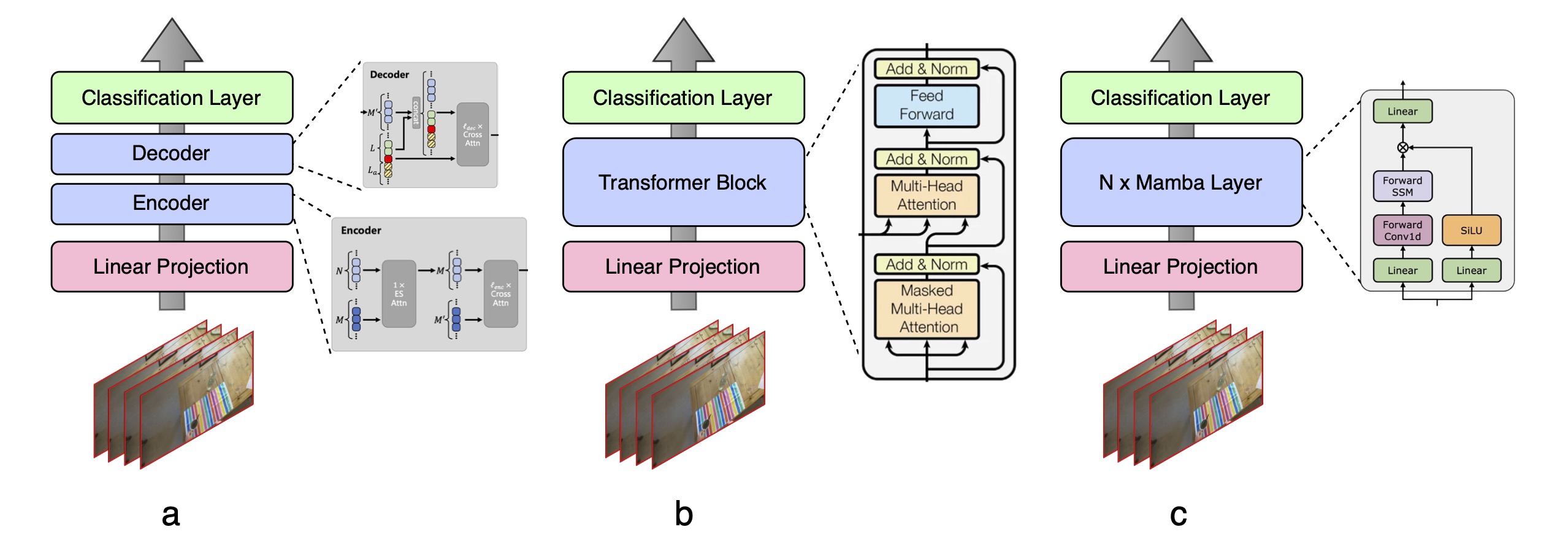}
    \caption{Models architecture overview. a) TeSTra uses an encoder for long-term memory and a decoder to combine long- and short-term information. b) Transformer processes only short-term memory using standard attention blocks. c) MAMBA-OTR employs one or more Mamba layers for efficient temporal modeling.}
    \label{fig:MAMBA-OTR}
\end{figure}

\subsection{Base Architectures}
All base architectures take as input visual features extracted from input frames and output an action prediction for each frame. Possible actions are ``take'', ``release'' or ``background''. Given this output, we extract $(\hat{c},\hat{t},s)$ tuples from each video.

\paragraph{\textbf{TeSTra}} We consider the TeSTra architecture~\cite{zhao2022real} as a state-of-the-art Transformer-based model optimized for online video processing.
TeSTra leverages a dual-memory structure, where an encoder builds a long-term memory from past observations, which is then fused with a short-term memory via cross-attention in a decoder to produce frame-level predictions. See \cref{fig:MAMBA-OTR}(a).

\paragraph{\textbf{Transformer}} We also consider a simplified version of TeSTra where the long-term memory module is removed, retaining only the self-attention mechanism over a short temporal window. This leads to a standard transformer module operating over the video in a sliding-window fashion. 
We train this model based on the TeSTra code-based, hence taking advantage of its data-loading and pre-processing pipeline. See \cref{fig:MAMBA-OTR}(b).


\paragraph{\textbf{Mamba}}
We finally consider a Mamba~\cite{gu2023mamba} architecture as a base-model. This architecture ingest linearly projected visual features extracted from frames and stacks different Mamba layers followed by a classification head. 
See \cref{fig:MAMBA-OTR}(c).


\subsection{Training Loss and Regularization Techniques}
We train our model using the following loss function:
\begin{equation}
    \mathcal{L} = \sum_i^m FL(h(x_i)) + \lambda R 
\end{equation}
where $FL$ is the focal loss calculated on the output of the model, given a specific dataset example $h(x_i)$ where $h$ is the used model, and $R$ is the regularization term. 

The focal loss~\cite{lin2017focal}, originally introduced for object detection, is a widely used technique for addressing class imbalance by down-weighting easy examples and focusing the learning on hard, misclassified ones. In our case, when each action is represented by a single positive frame among thousands of negatives, this imbalance is particularly severe. We show that the focal loss plays a crucial role in guiding the model to concentrate on these precise endpoints, rather than learning from potentially misleading earlier frames and overfitting to the majority class.

The regularization term $R$ is designed to reduce redundant detections around the same ground truth ``take'' or ``release'' action while maintaining model confidence. 
Specifically, we propose three different approaches to implement this regularization, which are described in the following: Entropy Minimization, Sliding Window Regularization, and Fixed Window Regularization. 

\paragraph{\textbf{{Entropy Minimization}}}
Entropy minimization offers a simple way to promote sparsity by encouraging confident predictions and suppressing uncertain ones. The loss is defined as:
\begin{equation}
\mathcal{R}_{entropy} = -\sum_{i} p_i \log(p_i)
\end{equation}
where \(p_i\) is the predicted probability at frame \(i\). While effective at reducing multiple low-confidence outputs, this approach lacks control over where predictions are concentrated in time.

\paragraph{\textbf{{Sliding Window}}}
To refine the prediction distribution, we use a sliding window approach. Given a window size \(w\), the loss sums the predictions within each window:
\begin{equation}
\mathcal{R}_{SW} = \sum_{f} \left( \sum_{i=f - \lfloor w/2 \rfloor}^{f + \lfloor w/2 \rfloor} p_i \right)
\end{equation}
where $f$ stands for each frame in the output vector.

\begin{figure}[t]
    \centering
    \includegraphics[width=\linewidth]{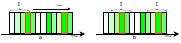}
    \caption{(a) Sliding Window: a window slides on all frames, encouraging sparse predictions. (b) Fixed Window: we place windows only around ground truth actions encouraging sparse predictions only in those parts of the video.}
    \label{fig:regularization}
\end{figure}

This encourages the model to output a single high-confidence prediction per ground truth by suppressing excessive activations within each segment, as shown in \cref{fig:regularization} (a). However, since the approach treats both background and action-relevant frames equally, it may penalize non-action regions.

\paragraph{\textbf{{Fixed Window}}}

To directly address the issue of multiple predictions around ground truth instances, we refine the sliding window approach by centering the window exclusively on the ground truth frames. Instead of applying the penalty over the entire sequence, we define a fixed window of size $w$ around each ground truth instance and minimize the sum of predictions within this region:
\begin{equation}
\mathcal{R}_{FW} = \sum_{g \in G} \left( \sum_{i=g - \lfloor w/2 \rfloor}^{g + \lfloor w/2 \rfloor} p_i \right)
\end{equation}
where $G$ represents the set of ground truth action-ending frames (See~\cref{fig:regularization} (b).

\section{Experiments}
In this section we consider the dataset introduced in \cite{catinello863online}, which is a curated subset of EPIC-KITCHENS-100, containing only two action classes: \textit{take} (e.g., ``get'', ``fetch'') and \textit{release} (e.g., ``put'', ``leave-on'').
Experiments aim to assess the different design choices leading to the definition of the final Mamba-OTR architecture. Due to the wide range of experimental settings, results are presented progressively, with only top-performing settings carried forward.
All models were trained using high-level features from \cite{shou2018online}, sampled at 4~fps, and evaluated using the mp-mAP metric, with numbers reported in percentage.

\subsection{Baseline Performance and Effect of Using the Focal Loss}
\cref{tab:focal_loss_results} reports the performance of the three base architectures when trained using standard cross entropy or focal loss.
The top section of the table reproduces the baseline models from \cite{catinello863online}: \textit{TeSTra} includes a 2-second long memory (8 frames) alongside short-term memory; \textit{Transformer} removes the long memory, using a single transformer block with short memory; and \textit{Mamba} replaces the transformer block with two Mamba layers. All models use 1 second (4 frames) for the short memory.\footnote{See~\cite{catinello863online} for more ablations on these models, for which we only report optimal configurations.} As can be observed, Transformer architectures struggle leveraging long-range information, with TeSTra and a standard Transformer performing on par. In these settings, Mamba already brings a better performance.

\begin{table}[t]
\setlength{\tabcolsep}{8pt} 
    \centering
    \resizebox{0.8\linewidth}{!}{
    \begin{tabular}{lccc}
        \hline
        \textbf{Model} & \textbf{Long Memory} & \textbf{Focal Loss} & \textbf{mp-mAP}$\uparrow$ \\
        \hline
        TeSTra & 8 frames & -- & 20.10 \\
        Transformer & -- & -- & 20.32 \\
        Mamba & -- & -- & 25.16 \\
        \hline
        TeSTra & 8 frames & \checkmark & \textbf{36.59} \\
        Transformer & -- & \checkmark & \textbf{38.48} \\
        Mamba & -- & \checkmark & \textbf{41.01} \\
        \hline
    \end{tabular}
    }
    \caption{Comparison of different models with and without Focal Loss.}
    \label{tab:focal_loss_results}
\end{table}

The lower section shows the same models trained with focal loss. The results demonstrate a substantial improvement across all models, with \textit{Mamba-OTR} achieving the best performance ($25.16 \rightarrow 41.01)$. These findings confirm the critical role of focal loss in handling the extreme class imbalance arising in this task. We use focal loss for training in all subsequent experiments.

\subsection{Number of Mamba layers and input frames during training}


\cref{tab:mamba_layer} presents the effect of varying the number of layers in the Mamba architecture. All models were trained with 4 frames (1 second) as input with the focal loss. Our results show that increasing the number of layers from 2 to 3 led to a small improvement in performance, from $41.01$ to $41.63$, while further increasing the number of layers to 4 resulted in a performance degradation, with the model achieving even worse results than the 2-layer configuration, all while significantly increasing the model size.
Since the configuration with 3 Mamba layers resulted in improved performance, it will be used in the subsequent experiments.

\begin{table}[t]
\setlength{\tabcolsep}{8pt} 
    \centering
    \begin{minipage}[t]{.49\linewidth}
    \centering{
    \centering
    \begin{tabular}{cc}
        \hline
        \textbf{\# Mamba layers} & \textbf{mp-mAP}$\uparrow$ \\
        \hline
        2 & 41.01 \\
        3 & \textbf{41.63} \\
        4 &  40.52\\
        \hline
    \end{tabular}
    \caption{Performance of Mamba when varying the number of layers.}
    \label{tab:mamba_layer}
    }
    \end{minipage}
    \hfill
    \begin{minipage}[t]{.49\linewidth}
    \centering{
        \centering
        \begin{tabular}{lc}
            \hline
            \textbf{Input frames} & \textbf{mp-mAP}$\uparrow$ \\
            \hline
            4 (1 second) & 41.63 \\
            8 (2 seconds) & 41.84 \\
            12 (3 seconds) & 42.47 \\
            20 (5 seconds) & \textbf{42.98} \\
            40 (10 seconds) &  41.77\\
            \hline
        \end{tabular}
        \caption{Performance of Mamba when varying the number of frames fed to the model during training.}
        \label{tab:memory}
        }
    \end{minipage}

\end{table}

\cref{tab:memory} demonstrates how varying the number of input frames impacts the performance of the Mamba model. As observed, increasing the number of input frames improves performance, with an increase from 4 frames (1s) to 20 frames (5s), where the mp-mAP rises from 41.63 to 42.98. However, when the number of input frames is increased to 40 (10s), the improvement over the 4-frames model is marginal and not as significant when compared to the 20-frames (5s) configuration. 
This is coherent with the observation that in the considered dataset each action lasts around $8$ frames ($\sim 2s$) in average. Thus, a window of $20$ frames allows to capture both the action and enough context on previous relevant interactions.

For the purposes of the subsequent experiments, we will use 20-frames model for subsequent experiments.

It is worth noting that, while processing longer sequences is challenging for transformer architectures, it is much more natural for Mamba models, due to their recurrent nature. Nevertheless, we see benefits in keeping the input size restricted to 20 frames (5s) at training time, while we will show benefits when extending this window at inference time, leveraging the generalization ability of the Mamba architecture.

\subsection{Regularization}
Given that the point-level mAP metric associates each ground truth instance with the closest prediction in time and penalize all other predictions made in its vicinity, the ideal model behavior is to produce a single, high-confidence prediction for each ground truth event. The regularization techniques we introduce aim to encourage this behavior during training by guiding the loss to penalize multiple nearby predictions appropriately.
Throughout all experiments, we used a regularization weight $\lambda = 0.01$. 

\begin{table}[t]
\setlength{\tabcolsep}{8pt} 
    \centering
    \begin{minipage}[t]{.45\linewidth}
    \centering{
    \centering
    \begin{tabular}{lc}
        \hline
        \textbf{Regularization} & \textbf{mp-mAP}$\uparrow$ \\
        \hline
        None &  42.98 \\
        Entropy &  42.70 \\
        Sliding Window  & 43.42 \\
        Fixed Window & \textbf{45.48}\\
        \hline
    \end{tabular}
    \caption{Performance of MAMBA-OTR with different regularization techniques. Both the window-based approach use a 4~frames window size.}
    \label{tab:regularization}
    }
    \end{minipage}
    \hfill
    \begin{minipage}[t]{.45\linewidth}
    \centering{
    \centering
    \begin{tabular}{lc}
        \hline
        \textbf{Window size} & \textbf{mp-mAP}$\uparrow$ \\
        \hline
        4 frames & \textbf{45.48} \\
        12 frames  & 44.05 \\
        20 frames  & 44.40 \\
        \hline
    \end{tabular}
    \caption{Performance of MAMBA-OTR with the Fixed Window regularization when varying the window size.}
    \label{tab:window_size}
    }
    \end{minipage}
\end{table}

\cref{tab:regularization} reports the impact of different regularization strategies on the Mamba architecture with 3 layers and a input frames length of 20 frames (5s). Notably, entropy regularization alone does not improve model performance. This is likely because it promotes general sparsity without enforcing any locality or temporal structure, thus failing to capture the intended prediction behavior.
Better results arise from the window-based regularization approaches, which operate in the neighborhood of ground truth annotations. The \textit{Sliding Window} technique yields a modest improvement of $0.44$ mp-mAP, increasing performance from $42.98$ to $43.42$. More significantly, the \textit{Fixed Window} method--where regularization is applied strictly around each positive (either take or release) ground truth annotation--produces a substantial improvement of $2.06$ over the \textit{Sliding Window} approach and $2.5$ over the baseline model, achieving a final mp-mAP of $45.48$.


\begin{table}[t]
    \centering
    \setlength{\tabcolsep}{8pt} 
    \resizebox{\linewidth}{!}{
    \begin{tabular}{lccccc}
        \hline
        \textbf{Model} & \textbf{Long memory} & \textbf{Short memory} & \textbf{Regularization} & \textbf{Window size} & \textbf{mp-mAP}$\uparrow$ \\
        \hline
        TeStra & 8 frames & 4 frames & -- & -- & 36.59 \\
        Transformer & -- & 4 frames & -- & -- & 38.48 \\
        \hline
        TeStra & 8 frames & 4 frames & Fixed Window & 4 frames & \textbf{37.06} \\
        Transformer & -- & 4 frames & Fixed Window & 4 frames & \textbf{38.96} \\
        \hline
    \end{tabular}
    }
    \caption{Improvements in performance brought by the Fixed Window regularization when used with TeSTra and Transformer.}
    \label{tab:regularization_transf}
\end{table}

\cref{tab:window_size} ablates the effectiveness of the best-performing regularization strategy, \textit{Fixed Window}, when considering different window sizes. 
While the performance differences are relatively small, it is worth noting that even the lowest-performing configuration (12 frames, 44.05 mp-mAP) still outperforms the \textit{Sliding Window} regularization from \cref{tab:regularization} by $0.63$ mp-mAP. With a 20 frame window, the improvement increases to $0.98$ mp-mAP over \textit{Sliding Window}. Nevertheless, the optimal result is obtained with a window size of 4 frames (1s), achieving an mp-mAP of $45.48$, consistent with the results previously reported in \cref{tab:regularization}. 
This is coherent with our previous observation that an take and release actions have a short duration, lasting in average $8$ frames.
Our fixed window regularization suppresses duplicate predictions around ground truth locations, encouraging the model to make more sparse and accurate predictions at test time, which results in an improved point level mAP.
We refer to this model as Mamba-OTR.

\cref{tab:regularization_transf} further shows the effect of the fixed window regularization technique on the two transformer-based models. Even if by small margins, regularization is beneficial with these architectures as well.

\begin{table}[t]
\setlength{\tabcolsep}{8pt} 
    \centering
    \resizebox{0.8 \linewidth}{!}{
    \begin{tabular}{lcccc}
        & \multicolumn{2}{c}{\textbf{Time $\downarrow$}} & \multicolumn{2}{c}{\textbf{mp-mAP$\uparrow$}} \\
        \hline
        \textbf{Model} & \textbf{Video} & \textbf{Frame} &\textbf{Sliding}$\uparrow$ & \textbf{Streaming}$\uparrow$ \\
        \hline
        Mamba-OTR & \textbf{0.14s} & \textbf{8ns} & \textbf{45.48} & 43.35 \\
        Transformer & \textbf{0.14s} & \textbf{8ns} & 38.96 & 0.04 \\
        \hline
    \end{tabular}
    }
    \caption{Inference performance of Mamba-OTR and Transformer when processing videos using a sliding window and in streaming settings at test time.}
    \label{tab:full_video}
\end{table}

\subsection{Extension to longer input sequences at test time}
Restricting input sequences to 20 frames (5s) at training time enables a controlled and regularized learning, leading to best results. Nevertheless, processing videos in short chunks at test time is impractical for recurrent architectures such as Mamba, leading to slow inference due to the need of re-setting the hidden state of the model at every chunk or, worse, re-processing the same frame multiple time when a sliding window scheme is considered.
In this section, we show that, while trained on fixed-length chunks, Mamba-OTR can generalize to longer sequences at inference time, achieving faster processing and even increased performance. On the contrary, transformer-based architectures do not possess this ability, likely due to their explicit modeling of temporal relations with positional embeddings.




\cref{tab:full_video} compares Mamba-OTR with the best-performing Transformer variant when tested using a sliding window approach (as in previous experiments) and in streaming mode, processing the full video one frame at a time, without relying on chunking or sliding window approaches.
Note that this latter setting allows to deploy models more naturally, without requiring storing a buffer of previous observations.
For each model, we also report mean processing time per video and per frame. We leave out the time required for feature extraction from this analysis.
Both models were trained using the focal loss and the \textit{Fixed Window} regularization strategy with a window size of 4 frames. 
The Transformer model is trained with an input size of $1$ second ($4$ frames), no long memory and with the fixed window regularization with a window size of 4 frames, which proved optimal among the Transformer variants. 
The Mamba-OTR configuration includes 3 Mamba layers, an input frames length of 20-frames (5s) and a fixed window regularization with a window size of 4 frames. 
We note that both architectures introduce minimal computational overhead beyond the backbone, both requiring only around 0.14 seconds per video and 8 nanoseconds for each frame.
mp-mAP values confirms the inability of the Transformer-based model to deal with the shift between training-time and test-time processing. Mamba-OTR is robust and retains almost the same performance in streaming mode, obtaining an mp-mAP of $43.35$, a small decrease from the original $45.48$.




\subsection{Qualitative Results}

\begin{figure}[t]
\centering
\begin{subfigure}[b]{\linewidth}
   \includegraphics[width=\linewidth]{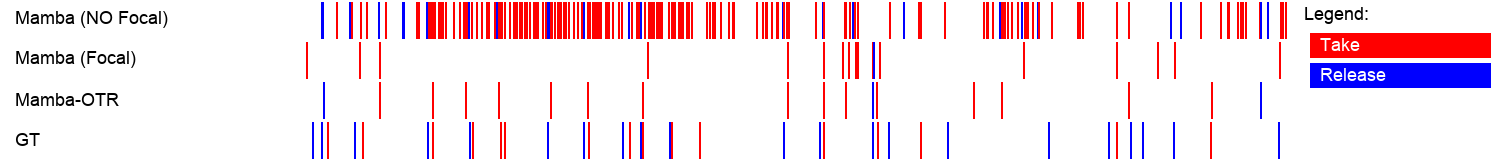}
\end{subfigure}
\par\bigskip
\begin{subfigure}[b]{\linewidth}
   \includegraphics[width=\linewidth]{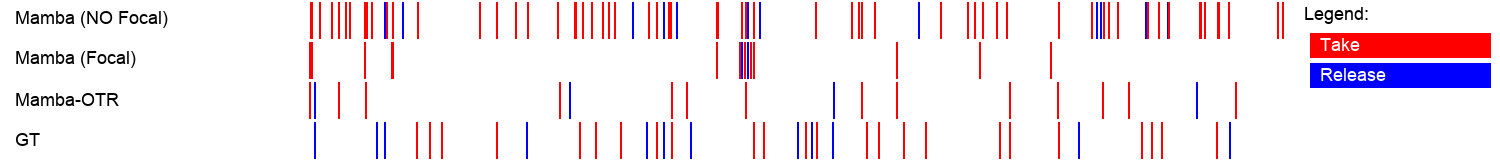}
\end{subfigure}
\caption{Qualitative comparison between different regularization techniques.}
\label{fig:qualitative}
\end{figure}

To further demonstrate the impact of Mamba-OTR on this task, we present a qualitative comparison in \cref{fig:qualitative}, where we evaluate Mamba without Focal Loss, Mamba with Focal Loss, and our proposed Mamba-OTR with regularization across two video examples exhibiting comparable levels of difficulty.
As illustrated, the model trained without Focal Loss produces numerous predictions without any clear correlation with the ground truth.
In contrast, both Mamba with Focal Loss and Mamba-OTR show improved behavior, with the latter displaying a marked advantage. 
Specifically, in the top example of \cref{fig:qualitative}, Mamba-OTR consistently generates a single, well-localized prediction for each ground truth event from the beginning to the midpoint of the video, a property not observed in the other models.
A similar trend is observed in the bottom example of \cref{fig:qualitative}, where Mamba with Focal Loss produces a dense cluster of predictions in the middle portion of the video, whereas Mamba-OTR successfully condenses this into a single take prediction.

\section{Conclusion}
We introduced Mamba-OTR, a Mamba-based model for the online detection of take and release actions in egocentric video. By combining focal loss with a fixed-window regularization approach, our method effectively addresses the challenges posed by class imbalance while aligning with the specific evaluation metric that penalizes multiple predictions around the same ground truth. Through extensive experiments, we shown that Mamba-OTR achieves state-of-the-art performance in terms of accuracy with a minimal overhead over the feature extractor, allowing to process full-length videos in real-time.

\section*{Acknowledgements}

This research has been funded by the European Union - Next Generation EU, Mission 4 Component 1 CUP E53D23008280006 - Project PRIN 2022 EXTRA-EYE, and FAIR – PNRR MUR Cod.
PE0000013 - CUP: E63C22001940006.

%
%
\bibliographystyle{splncs04}
\bibliography{main}
\end{document}